\documentclass[conference]{IEEEtran}
\IEEEoverridecommandlockouts
\usepackage{balance}
\usepackage{cite}
\usepackage{amsmath,amssymb,amsfonts}
\usepackage{graphicx}
\usepackage{textcomp}
\usepackage{xcolor}
\usepackage{url}
\usepackage{tabularx}
\usepackage{array}
\usepackage{algorithmic}
\usepackage{url}
\usepackage{hyperref}
\newcolumntype{C}[1]{>{\centering\arraybackslash}p{#1}}
\newcolumntype{L}[1]{>{\raggedright\arraybackslash}p{#1}}

\def\BibTeX{{\rm B\kern-.05em{\sc i\kern-.025em b}\kern-.08em
    T\kern-.1667em\lower.7ex\hbox{E}\kern-.125emX}}

\begin{document}

\title{FraudBench: Protocol-Sensitive Benchmarking of Adversarial Robustness for Financial Risk Assessment}

\author{
\IEEEauthorblockN{Xitong Zeng, Zhaoge Bi, Yitian Yang, and Huaming Chen}
\IEEEauthorblockA{
\textit{School of Electrical and Computer Engineering}\\
\textit{The University of Sydney}\\
Sydney, Australia
}
\and
\IEEEauthorblockN{Quan Z. Sheng}
\IEEEauthorblockA{
\textit{School of Computing}\\
\textit{Macquarie University}\\
Sydney, Australia
}
}

\maketitle

\begin{abstract}
Machine learning models are widely used in financial fraud and credit-risk detection, yet their adversarial robustness remains difficult to evaluate because financial tabular data involve domain-specific constraints, severe class imbalance, and asymmetric attacker capability. 
We argue that, in this setting, robustness is not only an attribute of the model, but also an attribute of the evaluation protocol. Different ways of enforcing constraints and capability can lead to substantially different robustness conclusions. This paper presents FraudBench, a protocol-sensitive benchmark for adversarial robustness evaluation in financial fraud and credit-risk detection. Rather than treating domain constraints as post-hoc validity checks, FraudBench evaluates the same dataset--model--attack--defence setting under three matched protocols: unconstrained attacks, post-hoc feasibility filtering, and deployment-aware constraint-integrated attacks. FraudBench covers four public financial datasets, and evaluates neural, tree-based, and ensemble models using three attack settings. Our results show that robustness conclusions are highly protocol-sensitive. On Lending Club Loan Data under the white-box setting, post-hoc filtering leaves only 3.7 feasible-flipped examples on average, whereas in-attack projection with attacker mutability masking produces 2,832.3 feasible-flipped examples under the same perturbation budget. The results on IEEE-CIS further show that feasibility and attacker capability are separate axes, while black-box evaluation shows that protocol choice can alter model-family rankings. These findings suggest that fraud robustness evaluation should report predictive degradation and attack feasibility jointly, and should incorporate domain constraints into attack generation rather than treating them as post-processing checks. The code repository is available at \href{https://github.com/iHaydenzZ/FraudBench}{{https://github.com/iHaydenzZ/FraudBench}}.

\end{abstract}


\section{Introduction}
Financial fraud is a sustained and growing problem. Global card payment
fraud losses reached \$33.41~billion in 2024, with card-not-present (CNP) transactions continuing to create substantial operational risk for issuers and merchants~\cite{lunghi2023realworld,nilson2026fraud}. Since fraudulent transactions are rare relative to legitimate activity,  manually reviewing every transaction is economically infeasible.
As a result, the screening of incoming transactions is overwhelmingly delegated to machine learning models that operate on financial tabular features.

Adversarial robustness in this setting differs from the image-classification setting that originally motivated much of adversarial machine learning \cite{goodfellow2014fgsm,madry2017pgd,croce2020apgd}. In image tasks, perturbations are commonly formalised as bounded changes within an $L_p$ ball. In tabular financial fraud detection, however, each feature has an operational meaning. A transaction amount must remain positive, categorical attributes must preserve valid category structure, and loan-related variables may be linked by arithmetic or contractual relationships. Moreover, attacker capability is asymmetric: an adversary may influence transaction-facing or self-reported fields, but cannot freely modify credit-bureau records, platform-internal risk scores, or historical account attributes. These properties make fraud robustness a domain-specific problem rather than a direct application of generic adversarial evaluation.

Another complication arises from the statistical properties of fraud data. Fraudulent cases are rare, and the positive-class rate can be far below one percent in credit-card transaction datasets~\cite{dalpozzolo2015ccfd}. Under such imbalance, accuracy is often misleading: a classifier may obtain high accuracy by predicting most records as legitimate while failing to identify fraudulent transactions. Financial fraud detection also relies heavily on tabular models such as gradient-boosted decision trees, including XGBoost and LightGBM, which remain strong baselines for structured data \cite{chen2016xgboost,ke2017lightgbm,grinsztajn2022tree}. A robustness benchmark for fraud detection should therefore not be restricted to neural models or accuracy-based evaluation. It should include production-relevant tree-based models, and prioritize metrics such as PR-AUC that reflect rare-class detection quality.

A separate challenge is the validity of adversarial examples. In ordinary adversarial classification, a successful attack is often defined only by whether the perturbed input changes the model prediction. In financial fraud detection, this criterion is not sufficient. A perturbed record that violates domain constraints is not a realistic adversarial example. Rather, it becomes an invalid financial record that can be rejected before model inference. Prior work on constrained tabular attacks has shown that tabular robustness requires constraints over feature ranges, categorical validity, feature mutability, and cross-feature relationships~\cite{ghamizi2020feature,chernikova2022fence,simonetto2022moeva,simonetto2024constrained}. Fraud-specific robustness evaluation must measure both predictive degradation and whether the generated adversarial records remain feasible.

One way to study such robustness is through standardised benchmarks. RobustBench\cite{croce2020robustbench} and ARES~\cite{dong2020ares} have demonstrated the value of fixed protocols and public comparability in image-domain adversarial robustness. In the tabular setting, TabularBench~\cite{simonetto2024tabularbench} provides an important benchmark with constrained attack infrastructure and a large-scale leaderboard for tabular deep learning models. Fraud detection has also been studied through public datasets and fraud-focused benchmarks covering credit-card transactions, online retail transactions, loan default, and broader abuse detection settings~\cite{ieeecis2019,sparkov2019,grover2022fdb}. In parallel, several studies have examined adversarial evasion against fraud detectors and attacker strategies in banking and payment systems \cite{carminati2020evasion,elawady2021reinforcement,cartella2021adversarial,fok2025foe}.

However, existing benchmarks still do not fully answer a central operational question for financial fraud detection: \textit{can an attacker, constrained by realistic feature mutability and domain validity requirements, evade the detector?} General adversarial-robustness benchmarks are mostly designed around image-domain threat models, while existing fraud benchmarks mainly focus on natural predictive performance or privacy-aware evaluation rather than adversarial robustness under domain constraints \cite{grover2022fdb,goldberg2025private}. Tabular robustness benchmarks provide valuable constraint-aware infrastructure, but their evaluation is not organised around the fraud-specific dimensions of severe class imbalance, production-relevant tree models, asymmetric attacker capability, and feasibility-aware attack success. In particular, post-hoc constraint filtering can measure how many generated attacks are infeasible, but it cannot determine how successful an attacker would be if domain constraints were enforced during attack generation. Thus, robustness conclusions may reflect the evaluation protocol rather than the intrinsic robustness of the model.

To address this gap, we present FraudBench, a protocol-sensitive adversarial robustness benchmark for tabular financial fraud and credit-risk detection. The key idea is to treat the evaluation protocol itself as a first-class experimental variable. FraudBench is designed to isolate how robustness conclusions change when the same experimental setting is evaluated under three protocols: unconstrained attack generation, post-hoc feasibility filtering, and deployment-aware constraint-integrated attack generation. The benchmark therefore treats protocol sensitivity as the main object of study, rather than reporting robustness as a single attack score in which domain constraints are ignored. This paper makes four contributions:
\begin{itemize}
    \item We introduce FraudBench, a reproducible benchmark for protocol-sensitive adversarial robustness evaluation in financial risk assessment, covering four public datasets, production-relevant model families, white-box and black-box attacks, and standard defense configurations.
    \item We formulate a deployment-aware fraud robustness evaluation protocol that distinguishes unconstrained attack success, post-hoc filterability, and constraint-aware attacker success through in-attack projection and attacker-capability masks.
    \item We show that protocol choice can materially change robustness conclusions. On LCLD, the feasible-flipped count changes from 3.7 under post-hoc filtering to 2,832.3 under projection plus mutability masking. On IEEE-CIS, projection increases feasible attacks but masking suppresses them, demonstrating that feasibility and capability are separate axes.
    \item We analyse model-family and defense conclusions under the same protocol view. Square Attack shows that protocol choice can change model selection across MLP, XGBoost, and ensemble models, while deployment-aware defense results show that adversarial training is the strongest evaluated defence and simple z-score input validation is not a reliable defence.
\end{itemize}

\section{Related Work}

\subsection{Adversarial Robustness Benchmarks}
Adversarial robustness has been extensively benchmarked in computer vision, where attacks are commonly formulated as bounded perturbations under an $L_p$ norm. Classical attacks such as FGSM and PGD established the standard evaluation setting, while later attack suites such as AutoAttack improved reliability by combining multiple strong attacks \cite{goodfellow2014fgsm,madry2017pgd,croce2020apgd}. RobustBench and ARES further showed the value of fixed protocols, strong attacks, and public comparability for robustness evaluation \cite{croce2020robustbench,dong2020ares}. However, these benchmarks are mainly designed for image-domain threat models. Their assumptions do not directly transfer to financial fraud detection, where input features have operational meanings and invalid perturbations can often be rejected before model inference.

\subsection{Constrained Tabular Adversarial Robustness}
Tabular adversarial robustness requires constraints that are absent or less explicit in image classification. Prior work has formalised tabular constraints in terms of feature immutability, domain validity, relational dependencies, and consistency between features \cite{ghamizi2020feature}. In financial data, such constraints are central: transaction amounts must remain positive, one-hot encoded categorical variables must remain valid, and loan-related fields may satisfy arithmetic relationships. Perturbations that violate these constraints are not realistic adversarial examples, because they correspond to invalid financial records.

Several attacks have been proposed for constrained tabular spaces. FENCE\cite{chernikova2022fence} combines gradient-based updates with constraint repair, while MOEVA\cite{simonetto2022moeva} searches for feasible adversarial examples through multi-objective evolutionary optimisation. CAPGD and CAA further extend projected-gradient-style attacks to constrained tabular data \cite{simonetto2024constrained}. TabularBench\cite{simonetto2024tabularbench} is the closest general-purpose benchmark, providing constraint specifications, constrained attack pipelines, and a large-scale leaderboard for tabular deep learning models. Nevertheless, its evaluation is not organised around fraud-specific deployment requirements such as class imbalance, tree-based production models, asymmetric attacker capability, and feasibility-aware attack success. FraudBench builds on this line of work by measuring how robustness conclusions change across evaluation protocols.

\subsection{Financial Fraud Detection Models and Benchmarks}
Machine learning has long been used in financial fraud detection, including credit-card fraud, online transaction fraud, loan default prediction, and synthetic transaction fraud settings \cite{dalpozzolo2015ccfd,ieeecis2019,sparkov2019,grover2022fdb}. A key challenge is \textit{severe class imbalance}: fraudulent cases are rare, and in some datasets the positive-class rate is far below one percent \cite{dalpozzolo2015ccfd}. Under such imbalance, accuracy can be misleading because a classifier may obtain high accuracy by predicting almost all records as legitimate. PR-AUC is therefore more informative for fraud detection.

Another important property is the continued strength of tree-based models. Gradient-boosted decision trees such as XGBoost and LightGBM remain strong baselines for structured tabular data and are widely used in practical risk-modelling pipelines \cite{chen2016xgboost,ke2017lightgbm,grinsztajn2022tree}. Deep tabular models have received substantial attention, but they do not consistently dominate tree-based methods on typical tabular datasets. A fraud-robustness benchmark that evaluates only differentiable neural networks therefore misses an important part of the deployed model landscape. Fraud-focused datasets and benchmarks have also supported progress in this area. Public resources such as CCFD, IEEE-CIS, Sparkov, and Lending Club data provide useful evaluation settings for different fraud and risk-modelling tasks \cite{dalpozzolo2015ccfd,ieeecis2019,sparkov2019}. The Fraud Dataset Benchmark unifies multiple fraud and abuse datasets under common loaders and evaluation procedures, while private fraud benchmarking has been studied in graph settings under differential privacy constraints \cite{grover2022fdb,goldberg2025private}. These resources standardise clean fraud detection evaluation, but do not systematically evaluate adversarial robustness under domain constraints, attack--defence configurations, and fraud-specific robustness metrics.

\subsection{Deployment Gaps in Fraud Robustness Evaluation}
Prior work on adversarial fraud detection has examined evasion attacks on banking models, reinforcement-learning-based attacker strategies, imbalanced tabular attacks, and transferable attacks against credit-card fraud detectors \cite{carminati2020evasion,elawady2021reinforcement,cartella2021adversarial,fok2025foe}. These studies show that fraud models can be vulnerable to adaptive manipulation, but the literature remains fragmented across datasets, model families, attacks, defences, and metrics.

This paper addresses four deployment gaps. First, fraud robustness should be evaluated with rare-class-sensitive metrics rather than accuracy alone. Second, production-relevant tree models should be evaluated alongside neural and ensemble models. Third, post-hoc filtering of invalid adversarial examples is not equivalent to generating feasible adversarial examples during the attack. Fourth, standard defences such as adversarial training, input validation, and ensembling should be compared under the same fraud-specific protocol. FraudBench is designed to measure these gaps directly by evaluating the same experimental setting under unconstrained attacks, post-hoc feasibility filtering, and deployment-aware constraint-integrated attacks.

\section{FraudBench Design}

\begin{figure*}[t]
\includegraphics[width=1\textwidth]{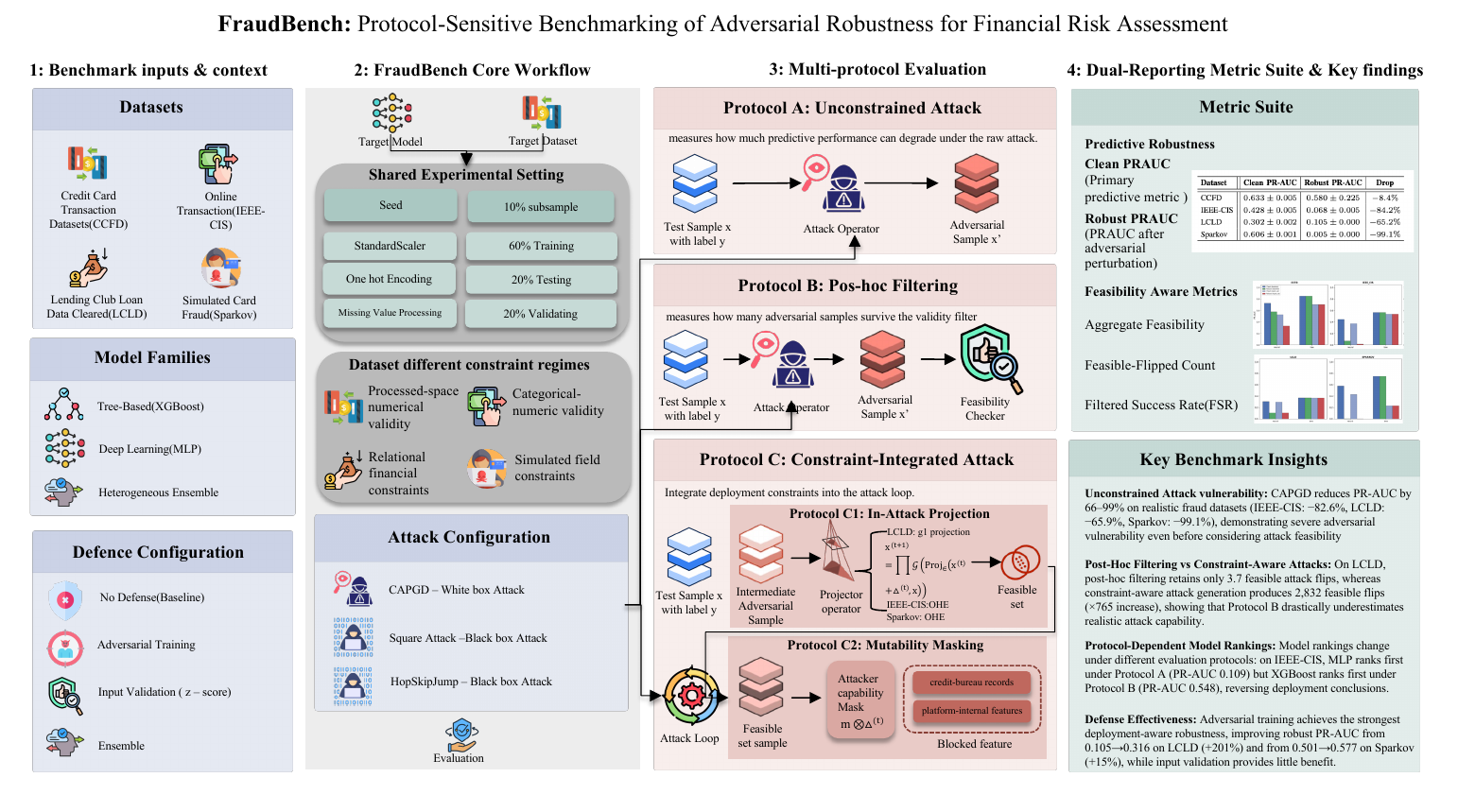}
\caption{Overview diagram of the FraudBench benchmark.}
\label{fig:overview}
\end{figure*}

\subsection{Benchmark Scope}
FraudBench is a config-driven benchmark for evaluating adversarial robustness in tabular financial fraud and credit-risk detection. While no particular fraud detector is introduced, FraudBench grounds the robustness conclusions with the evaluation protocol. It follows the domain relevant pattern with a unified experiment design, specifying a dataset, model family,  attack, defence configuration, perturbation budget, and random seed. The unified pipeline loads the dataset, applies a stratified split, preprocesses the features, trains the model, evaluates clean performance, constructs the constraint schema, generates adversarial examples, checks feasibility, and logs predictive and feasibility-aware metrics. Figure~\ref{fig:overview} provides an overview of the FraudBench pipeline.

Table~\ref{tab:coverage} summarises the experimental axes covered by FraudBench. The benchmark combines four public financial datasets, three model families, white-box and black-box attacks, four defence configurations, and three random seeds. This design allows the same setting to be evaluated under different robustness protocols rather than comparing unrelated model--dataset--attack combinations.

\begin{table}[!t]
\renewcommand{\arraystretch}{1.3}
\caption{FraudBench Experimental Coverage}
\label{tab:coverage}
\centering
\scriptsize
\setlength{\tabcolsep}{3pt}
\begin{tabular}{l||p{0.61\columnwidth}}
\hline
\bfseries Axis & \bfseries Coverage\\
\hline\hline
Datasets & CCFD, IEEE-CIS, LCLD, Sparkov\\
Models & MLP, XGBoost, heterogeneous ensemble\\
Attacks & CAPGD, Square Attack, HopSkipJump partial cross-check\\
Defences & None, adversarial training, input validation, ensemble\\
Seeds & 42, 123, 456\\
Primary metric & PR-AUC\\
Feasibility metrics & Aggregate feasibility, feasible-flipped count, filtered success rate\\
\hline
\end{tabular}
\end{table}

\subsection{Datasets and Constraint Structure}
Table~\ref{tab:datasets} summarises the four public datasets used in FraudBench,  covering credit-card fraud, online transaction fraud, loan default, and simulated transaction fraud. LCLD is a credit-risk/default task rather than a transaction-fraud dataset. It is included because it provides rich financial-domain constraints and a realistic setting in which borrower-controlled and institution-controlled fields differ. The selected datasets differ strongly in positive-class prevalence, feature semantics, and constraint structure, allowing the benchmark to test whether robustness conclusions hold across both highly imbalanced fraud tasks and a less imbalanced credit-risk task.

\begin{table}[!t]
\renewcommand{\arraystretch}{1.3}
\setlength{\tabcolsep}{3pt}
\caption{FraudBench Datasets}
\label{tab:datasets}
\centering
\scriptsize
\begin{tabular}{l||r r r r l}
\hline
\bfseries Dataset & \bfseries Samples & \bfseries Raw & \bfseries Proc. & \bfseries Pos. & \bfseries Scenario\\
\hline\hline
CCFD & 284,807 & 30 & 30 & 0.17\% & Card fraud\\
IEEE-CIS & 590,540 & 392 & 537 & 3.5\% & Online fraud\\
LCLD & 1,340,968 & 63 & 188 & 20.0\% & Loan default\\
Sparkov & 1,296,675 & 11 & 75 & 0.6\% & Simulated card fraud\\
\hline
\end{tabular}
\end{table}

The datasets are treated as static tabular binary classification tasks rather than temporal forecasting tasks. FraudBench first draws a stratified 10\% subsample for computational tractability and then applies a stratified 60/20/20 train/validation/test split. The cached split is reused across model, attack, and defence configurations within each seed. Numerical variables are standardised with \texttt{StandardScaler}, categorical variables are one-hot encoded, and missing values are imputed according to feature type.

The datasets provide different constraint regimes. CCFD is largely PCA-anonymised and therefore has no meaningful semantic constraint schema beyond processed-space numerical validity. It is used as a negative-control setting in which post-hoc filtering should not change protocol conclusions. IEEE-CIS contains one-hot validity constraints for categorical transaction fields such as \texttt{ProductCD}, \texttt{card4}, and \texttt{card6}, as well as non-negativity or positivity constraints for amount and count-like features. LCLD contains the richest explicit relational structure, including the loan instalment amortisation formula, inequalities such as \texttt{open\_acc} $\leq$ \texttt{total\_acc}, and one-hot validity for loan term. Sparkov contains simulated transaction fields with one-hot constraints for state, category, and gender, together with range constraints for transaction amount, city population, and merchant location.

\subsection{Models, Attacks, and Defences}
FraudBench evaluates three model families. The tree model XGBoost \cite{chen2016xgboost} is included because gradient-boosted trees remain a common production choice for tabular fraud detection. The neural model is a two-hidden-layer multilayer perceptron with ReLU activations, trained using class-weighted binary cross-entropy and the Adam optimiser. The heterogeneous ensemble model combines logistic regression, XGBoost, and the MLP through soft voting by averaging predicted fraud probabilities. The heterogeneous ensemble is treated as a model-family configuration and as a cross-model robustness comparison, rather than as a neural-model defence.

The primary white-box attack is Constrained Adaptive Projected Gradient Descent (CAPGD), following constrained tabular robustness work \cite{simonetto2024constrained}. In FraudBench, unmodified CAPGD is used as the Protocol A and Protocol B attack generator for differentiable neural evaluation. Protocol C variants extend the same attack loop with dataset-specific projection or masking operators. Unless otherwise specified, CAPGD uses an $L_\infty$ perturbation budget of $\epsilon=0.1$ in processed feature space and ten attack iterations.

FraudBench also includes Square Attack\cite{andriushchenko2020square} and HopSkipJump\cite{chen2020hopskipjump} as black-box companions. Square Attack is the cross-model black-box backbone because it is run over the four datasets and three model families with multi-seed coverage. HopSkipJump is included as a decision-based cross-check, but its coverage is partial due to computational cost. Since XGBoost is non-differentiable, gradient-based CAPGD results on the tree model are treated as clean-reference or no-op rows rather than valid white-box robustness estimates. Tree-model robustness is therefore read primarily from black-box attacks.

FraudBench evaluates four defence configurations: no defence, adversarial training, input validation, and heterogeneous ensembling. The no-defence setting serves as the baseline. Adversarial training augments neural-model training with CAPGD-generated adversarial examples and uses the same perturbation scale as evaluation \cite{tramer2017ensemble}. Input validation is implemented as z-score clipping based on training-set statistics: numerical values outside a three-standard-deviation range are clipped before prediction. The ensemble defence replaces the single classifier with the soft-voting ensemble described above.

\subsection{Protocol Coverage}
Table~\ref{tab:protocol_coverage} summarises the coverage of the protocol mechanisms. The table makes the benchmark scope explicit. Protocol C is implemented for the main constraint mechanisms needed by LCLD, IEEE-CIS, and Sparkov, while CCFD serves as a negative-control dataset because its PCA-anonymised features do not support a semantic constraint catalogue.

\begin{table}[!t]
\renewcommand{\arraystretch}{1.3}
\caption{Protocol Coverage Matrix}
\label{tab:protocol_coverage}
\centering
\scriptsize
\setlength{\tabcolsep}{2pt}
\resizebox{\columnwidth}{!}{
\begin{tabular}{l||l|l|c|c|c}
\hline
\bfseries Dataset & \bfseries Main Constraints & \bfseries C Operator & \bfseries CAPGD & \bfseries Square & \bfseries HSJ\\
\hline\hline
CCFD & None/PCA & None & A/B & A/B & Partial\\
IEEE-CIS & OHE, non-neg. & OHE+mask & A/B/C1/C2 & A/B & Partial\\
LCLD & $g_1$, inequalities, OHE & $g_1$+mask & A/B/C1/C2 & A/B & Partial\\
Sparkov & OHE, ranges & OHE+mask & A/B/C1/C2 & A/B & Partial\\
\hline
\end{tabular}
}
\end{table}

\section{Deployment-Aware Evaluation Protocol}

FraudBench evaluates adversarial robustness through three matched protocols. The protocols share the same dataset split, preprocessing pipeline, model, attack budget, and random seed. They differ only in how domain constraints and attacker capability are handled during evaluation. This design isolates the effect of the evaluation protocol itself.

\begin{table}[!t]
\renewcommand{\arraystretch}{1.3}
\setlength{\tabcolsep}{3pt}
\caption{Comparison of Evaluation Protocols}
\label{tab:evaluation_protocols}
\centering
\scriptsize
\begin{tabular}{c||c c p{0.48\columnwidth}}
\hline
\bfseries Protocol &
\bfseries \shortstack{Constraints\\in attack} &
\bfseries \shortstack{Post-hoc\\filtering} &
\bfseries Interpretation\\
\hline\hline
A & $\times$ & $\times$ & Unconstrained adversarial robustness\\
B & $\times$ & $\checkmark$ & Feasible subset after filtering unconstrained attacks\\
C & $\checkmark$ & $\checkmark$ & Constraint-aware attacker success under deployment conditions\\
\hline
\end{tabular}
\end{table}

\subsection{Protocol A: Unconstrained Attack}
Protocol A is the unconstrained adversarial baseline. Given a clean test sample $x$ with label $y$, the attack operator generates an adversarial sample $x'$ within the perturbation budget. No domain constraint is enforced during attack generation, and no infeasible sample is removed after the attack. Protocol A therefore measures how much predictive performance can degrade under the raw attack.

This protocol is useful as a stress test, but it is not a complete fraud-robustness estimate. In tabular financial data, unconstrained perturbations can easily produce invalid records. For example, a continuous attack may turn a one-hot categorical block into fractional values or break the consistency between a loan amount, interest rate, term, and instalment value. Such samples may still fool the model, but they are not necessarily valid financial records.

\subsection{Protocol B: Post-Hoc Filtering}
Protocol B uses the same adversarial samples generated by Protocol A, but applies a feasibility checker after attack generation. Let $\mathcal{G}(x')$ denote a constraint checker that returns one if $x'$ satisfies all implemented domain constraints and zero otherwise. Protocol B keeps only the feasible subset:
\begin{equation}
\mathcal{S}_{B} = \{x' : \mathcal{G}(x') = 1\}.
\label{eq:protocol_b_subset}
\end{equation}
Predictive metrics under Protocol B are then computed on the feasible adversarial subset.

Protocol B answers a specific question: among the adversarial examples produced by an unconstrained attack, how many survive the validity filter? It does not answer whether a constraint-aware attacker could generate feasible adversarial examples directly. Therefore, a low feasible-flipped count under Protocol B may reflect attack-generation inefficiency rather than genuine model robustness.

\subsection{Protocol C: Constraint-Aware Attack Generation}
Protocol C integrates deployment constraints into the attack loop. Instead of generating arbitrary adversarial samples and filtering them afterward, Protocol C repairs or restricts the intermediate adversarial sample after each attack step. In FraudBench, Protocol C is implemented through two mechanisms: in-attack projection and mutability masking. These mechanisms correspond to two different deployment requirements. Projection enforces feasibility, while masking enforces attacker capability.

\subsubsection{C1: In-Attack Projection}
C1 applies projection after each attack step. The attack first proposes an intermediate adversarial sample, and the projection operator maps selected feature groups back into the feasible set:
\begin{equation}
x^{(t+1)} = \Pi_{\mathcal{G}}\left(\mathrm{Proj}_{\epsilon}
\left(x^{(t)} + \Delta^{(t)}, x\right)\right),
\label{eq:c1_projection}
\end{equation}
where $\Delta^{(t)}$ is the attack update, $\mathrm{Proj}_{\epsilon}$ enforces the perturbation budget, and $\Pi_{\mathcal{G}}$ repairs the implemented domain constraints.

On LCLD, the main projection is the $g_1$ projection, which enforces the instalment amortisation constraint:
\begin{equation}
\mathrm{installment} =
\frac{\mathrm{loan\_amnt}\cdot r(1+r)^t}{(1+r)^t-1},
\label{eq:installment}
\end{equation}
where $r$ is the monthly interest rate and $t$ is the loan term in months. After each attack step, the perturbed \texttt{installment} value is overwritten by the value implied by the current \texttt{loan\_amnt}, \texttt{int\_rate}, and \texttt{term}. The attacker may still perturb loan-related variables, but the resulting record must remain consistent with the loan-payment formula.

On IEEE-CIS and Sparkov, the main projection restores one-hot encoding validity. For a categorical block $z=(z_1,\ldots,z_K)$, the projection is
\begin{equation}
\Pi_{\mathrm{OHE}}(z)_j =
\begin{cases}
1, & j = \arg\max_k z_k,\\
0, & \mathrm{otherwise}.
\end{cases}
\label{eq:ohe_projection}
\end{equation}
This prevents continuous attacks from producing invalid categorical mixtures, such as a card type that is partly Visa and partly Mastercard.

\subsubsection{C2: In-Attack Projection with Mutability Masking}
C2 extends C1 by adding an attacker-capability mask. A mutability mask $m \in \{0,1\}^d$ specifies which processed features the attacker is allowed to modify. The attack update is masked before the sample is updated:
\begin{equation}
\Delta^{(t)} \leftarrow m \odot \Delta^{(t)}.
\label{eq:mutability_mask}
\end{equation}
Immutable features therefore remain unchanged throughout the attack. The mask is derived from a raw-feature threat model and then mapped into processed feature space after preprocessing. If a raw categorical feature is mutable, all one-hot dimensions derived from it are mutable. If the raw feature is immutable, the corresponding processed dimensions are frozen. This captures the asymmetric attacker capability in fraud settings. A borrower may influence self-reported fields such as loan amount, income, purpose, or term, but cannot directly edit credit-bureau records or platform-internal features. Similarly, an online transaction attacker may control transaction-facing fields but not opaque internal risk features.

\subsection{Protocol Execution}
Fig.~\ref{alg:fraudbench_protocols} summarises the execution of the three protocols. Protocol A returns the unconstrained adversarial sample. Protocol B uses the same sample but evaluates only the feasible subset. Protocol C1 inserts projection into the attack loop, and Protocol C2 further applies mutability masking before each projected update.

\begin{figure}[!ht]
\centering
\footnotesize
\begin{minipage}{0.98\columnwidth}
\renewcommand{\algorithmicrequire}{\textbf{Input:}}
\renewcommand{\algorithmicensure}{\textbf{Output:}}
\begin{algorithmic}[1]
\REQUIRE Test sample $x$, label $y$, trained model $f$, attack operator $\mathcal{A}$, constraint checker $\mathcal{G}$, projection operator $\Pi_{\mathcal{G}}$, mutability mask $m$, protocol $p \in \{A,B,C1,C2\}$
\ENSURE Adversarial sample $x'$, prediction flip indicator $\mathrm{Flip}$, feasibility indicator $\mathrm{Feas}$

\STATE Initialise $x^{(0)} \leftarrow x$
\FOR{$t = 0$ to $T-1$}
    \STATE Generate attack update $\Delta^{(t)} \leftarrow \mathcal{A}(f, x^{(t)}, y)$
    \IF{$p = C2$}
        \STATE Apply mutability mask: $\Delta^{(t)} \leftarrow m \odot \Delta^{(t)}$
    \ENDIF
    \STATE Update sample: $\tilde{x}^{(t+1)} \leftarrow x^{(t)} + \Delta^{(t)}$
    \STATE Project to budget: $\tilde{x}^{(t+1)} \leftarrow \mathrm{Proj}_{\epsilon}(\tilde{x}^{(t+1)}, x)$
    \IF{$p = C1$ or $p = C2$}
        \STATE Apply in-attack projection: $x^{(t+1)} \leftarrow \Pi_{\mathcal{G}}(\tilde{x}^{(t+1)})$
    \ELSE
        \STATE Set $x^{(t+1)} \leftarrow \tilde{x}^{(t+1)}$
    \ENDIF
\ENDFOR

\STATE Set $x' \leftarrow x^{(T)}$
\STATE Compute prediction flip: $\mathrm{Flip} \leftarrow \mathbb{I}\{ f(x') \neq f(x) \}$
\STATE Compute feasibility: $\mathrm{Feas} \leftarrow \mathcal{G}(x')$

\IF{$p = B$}
    \STATE Keep $x'$ only if $\mathrm{Feas}=1$. Otherwise discard it from the feasible subset
\ENDIF

\STATE \textbf{return} $x'$, $\mathrm{Flip}$, $\mathrm{Feas}$
\end{algorithmic}
\end{minipage}
\caption{FraudBench evaluation protocols. Protocol A runs an unconstrained attack. Protocol B applies post-hoc filtering to unconstrained attacks. Protocol C1 applies in-attack projection after each attack step. Protocol C2 combines in-attack projection with attacker mutability masking.}
\label{alg:fraudbench_protocols}
\end{figure}

\subsection{Metrics}
FraudBench reports both predictive and feasibility-aware metrics. Predictive metrics include clean PR-AUC and robust PR-AUC. PR-AUC is the primary predictive metric because fraud datasets are often severely imbalanced, making accuracy unreliable as a ranking measure. Robust PR-AUC is computed on adversarially perturbed inputs.

Feasibility-aware metrics include aggregate feasibility, feasible-flipped counts, and filtered success rate. A flipped example is an input whose prediction changes after attack. A feasible-flipped example is both prediction-flipped and valid under all domain constraints. The filtered success rate is defined as
\begin{equation}
\mathrm{FSR} =
\frac{
\#\{\mathrm{flipped}\ \cap\ \mathrm{feasible}\}
}{
\#\{\mathrm{flipped}\}
}.
\label{eq:fsr}
\end{equation}
This metric is not the positive-class ratio. It is the proportion of successful prediction flips that also satisfy all domain constraints.

FraudBench reports feasibility metrics alongside PR-AUC because the two capture different aspects of robustness. PR-AUC measures ranking degradation under attack, whereas feasible-flipped count measures how many successful attacks remain plausible under the dataset's business rules and attacker-capability assumptions. A model can show severe robust-PR-AUC degradation even when most attacks are infeasible, or it can appear safe under post-hoc filtering while remaining vulnerable to an attacker that generates feasible adversarial records directly.

\section{Experimental Setup}
All experiments use the cached stratified subsample and split described above. Unless otherwise specified, the main CAPGD setting uses $L_\infty$ perturbations with $\epsilon=0.1$ in processed feature space and ten attack steps. This budget should be interpreted as a first-order robustness stress test in standardised feature space rather than as a complete economic attack-cost model. Future work should replace or complement this threat model with monetary, profit-based, and attacker-cost-aware perturbation budgets.

Experiments are replicated over seeds 42, 123, and 456. Results are logged to CSV registries with dataset, model, defence, attack, protocol, seed, perturbation budget, clean predictive metrics, adversarial predictive metrics, feasibility metrics, failed constraint labels, and runtime. The table-level artifact used for this manuscript consists of the CAPGD grid registry and the Square Attack model-family registry, which are sufficient to verify the numerical aggregates reported in the tables.

\section{Experimental Results}

\subsection{Unconstrained Attack Vulnerability}
Table~\ref{tab:mvb} reports the headline neural-model results without defence. CAPGD causes large degradation on IEEE-CIS, LCLD, and Sparkov, while CCFD is less affected but exhibits high seed variance due to its extremely small positive class.

\begin{table}[!t]
\renewcommand{\arraystretch}{1.3}
\caption{Clean and CAPGD-Robust PR-AUC on the Neural Model Without Defence}
\label{tab:mvb}
\centering
\small
\setlength{\tabcolsep}{3pt}
\begin{tabular}{l||c|c|c}
\hline
\bfseries Dataset & \bfseries Clean PR-AUC & \bfseries Robust PR-AUC & \bfseries Drop\\
\hline\hline
CCFD & $0.683\pm0.189$ & $0.598\pm0.262$ & $-12.5\%$\\
IEEE-CIS & $0.433\pm0.033$ & $0.075\pm0.013$ & $-82.6\%$\\
LCLD & $0.308\pm0.003$ & $0.105\pm0.000$ & $-65.9\%$\\
Sparkov & $0.626\pm0.033$ & $0.005\pm0.000$ & $-99.1\%$\\
\hline
\end{tabular}
\end{table}

These results support the use of PR-AUC as the primary predictive metric. Since fraud positives are rare, accuracy can remain high even when the model fails to rank fraudulent records above legitimate ones. However, PR-AUC alone is insufficient for robustness evaluation because it measures ranking degradation but not whether the adversarial records remain feasible.

\subsection{Post-Hoc Filtering versus Constraint-Aware Attack Generation}

\begin{figure*}[!ht]
\centering
\includegraphics[width=0.5\textwidth]{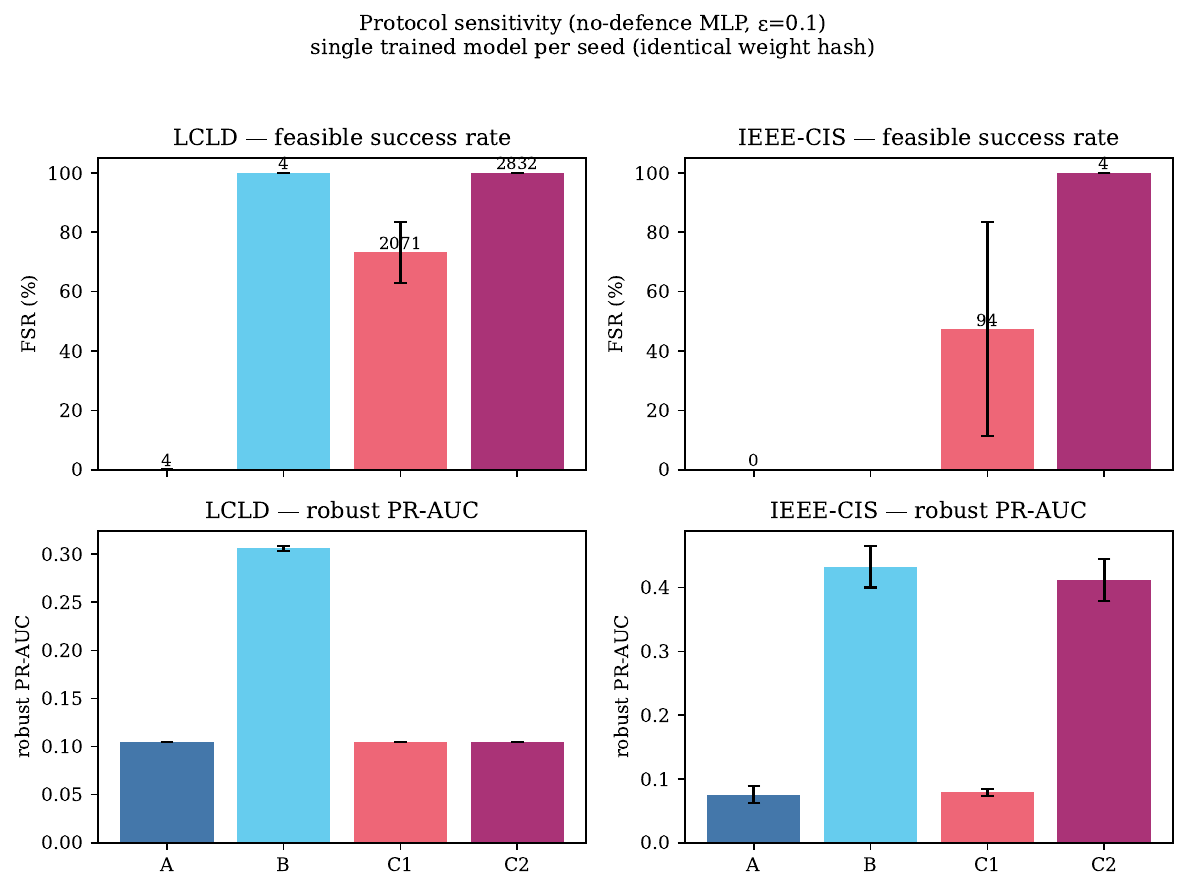}
\caption{Protocol sensitivity under CAPGD at $\epsilon=0.1$ on the no-defence MLP. The top panels show filtered success rate (FSR), with the annotated numbers indicating feasible-flipped counts. The bottom panels show robust PR-AUC. On LCLD, Protocol B retains only a few feasible-flipped attacks from the unconstrained attack, while Protocols C1 and C2 generate feasible attacks directly. On IEEE-CIS, C1 increases feasible attack success through one-hot projection, whereas C2 shows the suppressive effect of mutability masking.}
\label{fig:protocol_core}
\end{figure*}

Table~\ref{tab:capgd_protocol_comparison} compares the FraudBench protocols under CAPGD at $\epsilon=0.1$ with no defence. Protocol B PR-AUC is not directly comparable to Protocol A PR-AUC as an intrinsic robustness score because infeasible adversarial samples are removed before evaluation. Its role is to quantify the filterability of unconstrained attacks. C1 denotes in-attack projection, and C2 denotes in-attack projection with mutability masking. The visualisation is shown in Figure~\ref{fig:protocol_core}.

\begin{table}[!t]
\renewcommand{\arraystretch}{1.2}
\caption{CAPGD Protocol Comparison at $\epsilon=0.1$ with No Defence}
\label{tab:capgd_protocol_comparison}
\centering
\scriptsize
\setlength{\tabcolsep}{1pt}
\resizebox{\columnwidth}{!}{
\begin{tabular}{l||c|c|c|c|c}
\hline
\bfseries Dataset & \bfseries Prot. & \bfseries Robust PR-AUC & \bfseries Agg. Feas. & \bfseries Feas.-Flip. & \bfseries FSR\\
\hline\hline
CCFD & A & 0.598 $\pm$ 0.262 & 1.000 & 0.0 & --\\
CCFD & B & 0.598 $\pm$ 0.262 & 1.000 & 0.0 & --\\
\hline
IEEE-CIS & A & 0.075 $\pm$ 0.013 & 0.000 & 0.0 & 0.000\\
IEEE-CIS & B & 0.433 $\pm$ 0.033 & 1.000 & 0.0 & --\\
IEEE-CIS & C1 & 0.079 $\pm$ 0.005 & 0.453 & 93.7 & 0.474\\
IEEE-CIS & C2 & 0.412 $\pm$ 0.033 & 1.000 & 4.3 & 1.000\\
\hline
LCLD & A & 0.105 $\pm$ 0.000 & 0.001 & 3.7 & 0.001\\
LCLD & B & 0.306 $\pm$ 0.003 & 0.998 & 3.7 & 1.000\\
LCLD & C1 & 0.105 $\pm$ 0.000 & 0.776 & 2,071.0 & 0.732\\
LCLD & C2 & 0.105 $\pm$ 0.000 & 1.000 & 2,832.3 & 1.000\\
\hline
Sparkov & A & 0.005 $\pm$ 0.000 & 0.000 & 0.0 & 0.000\\
Sparkov & B & 0.626 $\pm$ 0.033 & 1.000 & 0.0 & --\\
Sparkov & C1 & 0.440 $\pm$ 0.040 & 1.000 & 17.0 & 1.000\\
Sparkov & C2 & 0.501 $\pm$ 0.033 & 1.000 & 13.0 & 1.000\\
\hline
\end{tabular}
}
\end{table}

On CCFD, Protocol A and Protocol B are identical because no semantic constraint catalogue is defined. This makes CCFD a negative-control case: when there are no domain constraints, post-hoc filtering does not change the result.

On LCLD, Protocol A reaches robust PR-AUC $0.105 \pm 0.000$, but only 3.7 flipped examples remain feasible. Protocol B reports a higher robust PR-AUC because infeasible adversarial records are removed before evaluation. This should be interpreted as attack filterability, not intrinsic model robustness. In contrast, C1 raises the feasible-flipped count to 2,071.0 by integrating the instalment formula during attack generation, and C2 further raises it to 2,832.3 with full feasibility. Thus, post-hoc filtering retains only a tiny fraction of the attacks that a constraint-aware attacker can generate directly.

IEEE-CIS and Sparkov show the same qualitative issue through categorical validity constraints. On IEEE-CIS, one-hot projection in C1 raises the feasible-flipped count from 0.0 to 93.7, while adding a conservative mutability mask in C2 reduces the count to 4.3. On Sparkov, Protocol A produces near-zero robust PR-AUC but zero feasible-flipped attacks, whereas C1 and C2 produce feasible-flipped attacks directly. These results show that one-hot validity is not a minor post-processing detail. Instead, it changes the operational meaning of attack success.

\subsection{Protocol-Dependent Model-Family Rankings}
\begin{figure*}[!ht]
\centering
\includegraphics[width=0.9\textwidth]{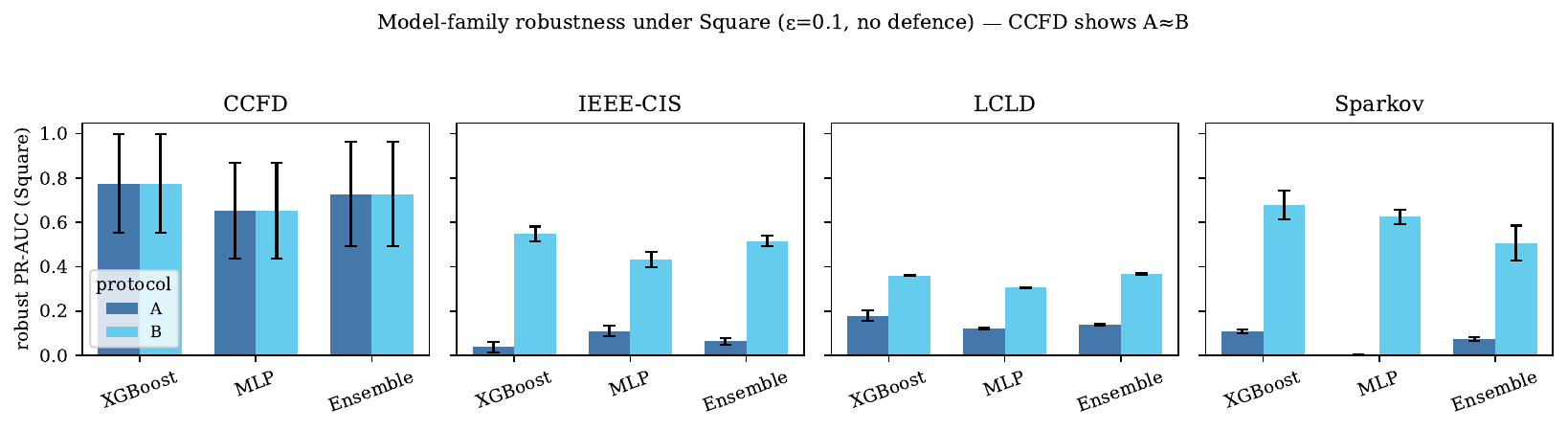}
\caption{Model-family robustness under Square Attack at $\epsilon=0.1$ with no defence. Bars show robust PR-AUC under Protocol A (unconstrained attack) and Protocol B (post-hoc filtering), with error bars indicating one standard deviation across seeds. CCFD shows almost identical Protocol-A and Protocol-B results because no semantic constraint catalogue is defined. In contrast, IEEE-CIS, LCLD, and Sparkov show large protocol-dependent shifts, demonstrating that post-hoc filtering can alter not only metric values but also model-family rankings.}
\label{fig:model_family_square}
\end{figure*}

Table~\ref{tab:square_family_results} reports Square Attack results across datasets and model families. These results are also visualised by Figure~\ref{fig:model_family_square}. Since Square Attack does not require gradient access, it is the main cross-model black-box attack and is especially important for evaluating XGBoost and the heterogeneous ensemble.

\begin{table}[!ht]
\renewcommand{\arraystretch}{1.3}
\caption{Square Attack results across datasets and model families. A denotes unconstrained attack, and B denotes post-hoc filtering.}
\label{tab:square_family_results}
\centering
\scriptsize
\setlength{\tabcolsep}{2pt}
\resizebox{\columnwidth}{!}{
\begin{tabular}{l||l|c|c|c|c|c}
\hline
\bfseries Dataset & \bfseries Model & \bfseries A PR-AUC & \bfseries B PR-AUC & \bfseries A Feas. & \bfseries A Flip. & \bfseries Time\\
\hline\hline
CCFD & MLP & 0.651 $\pm$ 0.216 & 0.651 $\pm$ 0.216 & 1.000 & 0.0 & 8.1\\
CCFD & XGBoost & 0.774 $\pm$ 0.223 & 0.774 $\pm$ 0.223 & 1.000 & 0.7 & 32.1\\
CCFD & Ensemble & 0.728 $\pm$ 0.237 & 0.728 $\pm$ 0.237 & 1.000 & 0.7 & 53.2\\
\hline
IEEE-CIS & MLP & 0.109 $\pm$ 0.024 & 0.432 $\pm$ 0.034 & 0.218 & 192.0 & 29.5\\
IEEE-CIS & XGBoost & 0.037 $\pm$ 0.024 & 0.548 $\pm$ 0.033 & 0.032 & 109.3 & 623.4\\
IEEE-CIS & Ensemble & 0.063 $\pm$ 0.016 & 0.516 $\pm$ 0.025 & 0.125 & 213.7 & 783.2\\
\hline
LCLD & MLP & 0.121 $\pm$ 0.003 & 0.305 $\pm$ 0.002 & 0.389 & 2,797.3 & 29.2\\
LCLD & XGBoost & 0.179 $\pm$ 0.022 & 0.360 $\pm$ 0.002 & 0.243 & 290.3 & 540.3\\
LCLD & Ensemble & 0.139 $\pm$ 0.005 & 0.367 $\pm$ 0.003 & 0.380 & 2,836.3 & 477.7\\
\hline
Sparkov & MLP & 0.003 $\pm$ 0.000 & 0.625 $\pm$ 0.033 & 0.440 & 163.0 & 52.7\\
Sparkov & XGBoost & 0.109 $\pm$ 0.009 & 0.678 $\pm$ 0.063 & 0.195 & 102.7 & 369.8\\
Sparkov & Ensemble & 0.074 $\pm$ 0.010 & 0.507 $\pm$ 0.079 & 0.014 & 76.3 & 526.8\\
\hline
\end{tabular}
}
\end{table}

\begin{table}[!ht]
\renewcommand{\arraystretch}{1.2}
\caption{Partial HopSkipJump Cross-Check on XGBoost. HopSkipJump is used only as a decision-based black-box cross-check on the tree model. Seed counts are shown in parentheses where fewer than three seeds were run.}
\label{tab:hsj_xgboost_results}
\centering
\scriptsize
\setlength{\tabcolsep}{2pt}
\begin{tabular}{l||c}
\hline
\bfseries Dataset & \bfseries HopSkipJump Robust PR-AUC\\
\hline\hline
CCFD & $0.002 \pm 0.001$\\
IEEE-CIS & $0.060 \pm 0.042$ (2 seeds)\\
LCLD & $0.185$ (1 seed)\\
Sparkov & --\\
\hline
\end{tabular}
\end{table}

The key information in Table~\ref{tab:square_family_results} is not only the robust PR-AUC value, but also the model-family conclusion induced by the protocol. Table~\ref{tab:model_ranking_change} summarises the best model under Protocols A and B. On IEEE-CIS, the most robust model under Protocol A is MLP, whereas Protocol B ranks XGBoost first. The top and bottom rankings are reversed. On LCLD, the best model changes from XGBoost under Protocol A to the ensemble under Protocol B. On Sparkov, XGBoost remains first but the lower-rank ordering changes. Thus, protocol choice can affect model selection, not merely the numerical value of a robustness metric. Table~\ref{tab:hsj_xgboost_results} reports HopSkipJump as a partial decision-based cross-check on XGBoost only because of the computational cost. As its coverage is incomplete, HopSkipJump is not used for the main model-family ranking analysis. Square Attack remains the primary black-box attack for cross-model comparison.

\begin{table}[!t]
\renewcommand{\arraystretch}{1.3}
\caption{Protocol-Induced Model-Family Ranking Changes under Square Attack}
\label{tab:model_ranking_change}
\centering
\scriptsize
\setlength{\tabcolsep}{5pt}
\begin{tabular}{l||l|l|l}
\hline
\bfseries Dataset & \bfseries Best A & \bfseries Best B & \bfseries Change\\
\hline\hline
CCFD & XGBoost & XGBoost & No top change\\
IEEE-CIS & MLP & XGBoost & Top--bottom reversal\\
LCLD & XGBoost & Ensemble & Top change\\
Sparkov & XGBoost & XGBoost & Lower-rank change\\
\hline
\end{tabular}
\end{table}

\subsection{Deployment-Aware Defences}

Table~\ref{tab:defence_results} reports robust PR-AUC for the evaluated defence settings under the deployment protocol used for each dataset. CCFD uses Protocol A because no semantic constraints are defined. IEEE-CIS, LCLD, and Sparkov use Protocol C2. The None, adversarial-training, and z-score input-validation columns are evaluated on the neural model, while the ensemble column is included as a cross-model comparison. This reporting choice aligns the defence comparison with the paper's central claim: robustness conclusions should be compared under the same deployment-aware protocol used for adversarial evaluation.

\begin{table}[!ht]
\renewcommand{\arraystretch}{1.3}
\caption{Deployment-Aware Defence Robust PR-AUC with Ensemble Comparison}
\label{tab:defence_results}
\centering
\scriptsize
\setlength{\tabcolsep}{2pt}
\resizebox{\columnwidth}{!}{
\begin{tabular}{l||c|c|c|c|c}
\hline
\bfseries Dataset & \bfseries Prot. & \bfseries None & \bfseries Adv. Train & \bfseries Z Input Val. & \bfseries Ensemble\\
\hline\hline
CCFD & A & $0.598 \pm 0.262$ & $0.580 \pm 0.225$ & $0.572 \pm 0.300$ & $0.794 \pm 0.074$\\
IEEE-CIS & C2 & $0.412 \pm 0.033$ & $0.453 \pm 0.023$ & $0.367 \pm 0.034$ & $0.030 \pm 0.008$\\
LCLD & C2 & $0.105 \pm 0.000$ & $0.316 \pm 0.004$ & $0.105 \pm 0.000$ & $0.108 \pm 0.001$\\
Sparkov & C2 & $0.501 \pm 0.033$ & $0.577 \pm 0.052$ & $0.326 \pm 0.013$ & $0.093 \pm 0.015$\\
\hline
\end{tabular}
}
\end{table}

Adversarial training is the strongest neural-model defence on the constrained datasets under Protocol C2, improving robust PR-AUC on IEEE-CIS, LCLD, and Sparkov. On CCFD, where no semantic constraint catalogue is defined and Protocol A is used, the ranking is less stable. The ensemble obtains the highest mean robust PR-AUC, while adversarial training does not improve over the no-defence baseline. Z-score input validation provides no reliable robust-PR-AUC benefit under the tested setting and is consistently weaker than adversarial training on the constrained datasets. The ensemble column should be interpreted as a cross-model comparison rather than a neural-model defence: it is beneficial on CCFD but weak on the constrained datasets, especially IEEE-CIS and Sparkov.

\section{Discussion}

\subsection{Protocol Sensitivity}
The main empirical message is that robustness conclusions on financial fraud and credit-risk data are protocol-sensitive. Protocol B measures how many outputs of a particular unconstrained attack survive a feasibility filter. If very few survive, the attack may simply be inefficient at producing valid records. That does not imply that the model is safe against an attacker who respects constraints during generation.

This is clearest on LCLD. Protocol A and Protocol B leave only 3.7 feasible-flipped examples, while Protocol C2 produces 2,832.3 feasible-flipped examples under the same perturbation budget. The number of realistic successful attacks is therefore determined by the evaluation protocol, not only by the model. This supports the central FraudBench claim that fraud robustness must be evaluated with in-attack constraint integration, not only post-hoc filtering.

\subsection{Feasibility and Capability Are Separate Axes}
The IEEE-CIS results show that feasibility and attacker capability should not be collapsed into one metric. One-hot projection makes categorical records feasible and raises feasible-flipped counts from 0.0 to 93.7. However, adding a mutability mask reduces the feasible-flipped count to 4.3 because the realistic attacker loses access to many predictive dimensions. This is the opposite of LCLD, where adding a mask after projection increases the feasible-flipped count.

The design of asymmetry in FraudBench is meant to be useful rather than problematic, which shows that realistic fraud robustness depends on the overlap between mutable features and predictive features. If the attacker can modify features that the model relies on, as in LCLD, constraint-aware attacks remain dangerous. If the predictive signal lies in immutable or institution-controlled features, as in IEEE-CIS, capability constraints can substantially reduce attacker success.

\subsection{Dual Reporting is Necessary}
The experiments show that no single metric is sufficient.
PR-AUC captures ranking degradation on imbalanced data, but does not indicate whether adversarial records are feasible.
Feasibility captures constraint validity, but does not measure
predictive degradation. Feasible-flipped count captures the
intersection of both: the number of attacks that both fool the
model and satisfy all constraints.

FraudBench therefore recommends reporting clean PR-AUC, robust PR-AUC, aggregate feasibility, feasible-flipped count, and filtered success rate together. The Square Attack results reinforce this need: several cells have very low robust PR-AUC under Protocol A but much higher Protocol-B robust PR-AUC after filtering. Without feasibility-aware metrics, these differences would be easy to misread as model robustness rather than attack filterability.

\subsection{Relation to TabularBench}
FraudBench should not be read as a replacement for TabularBench. TabularBench provides broad coverage of tabular deep learning architectures and constrained attack protocols \cite{simonetto2024constrained}. FraudBench narrows the scope to financial fraud and credit-risk detection and studies the consequences of severe class imbalance, production-relevant tree models, domain-specific constraints, and feasibility-aware attacker success. The benchmark therefore answers a different application question: whether models used in fraud detection remain reliable under perturbations that are both adversarial and valid under financial-domain constraints.

The results show why this application-specific framing matters. Under Protocol B, post-hoc filtering can make an attack look weak because most generated records are invalid. Under Protocol C, the attacker generates valid records during the attack process. On LCLD, this distinction changes the feasible-flipped count from 3.7 to 2,832.3. This is not a small metric adjustment. It changes the interpretation of model safety.

\section{Limitations and Responsible Use}
FraudBench has several limitations. The main experiments use only three seeds, which may be insufficient for highly imbalanced datasets such as CCFD. The $L_\infty$ budget in processed space is a standardised stress test rather than a monetary or attacker-cost-aware threat model. CAPGD is not directly applicable to XGBoost, so tree-model robustness should be read mainly from black-box attacks such as Square Attack and HopSkipJump. Broader comparison with CAA and MOEVA, as well as more datasets and projection operators, remains future work. FraudBench is intended for defensive evaluation, model auditing, and benchmark comparison, not as operational guidance for committing fraud.

\section{Conclusion}
This paper has presented FraudBench, a protocol-sensitive benchmark for adversarial robustness in tabular financial fraud and credit-risk detection. FraudBench combines public financial datasets, production-relevant model families, white-box and black-box attacks, PR-AUC-centred evaluation, and feasibility-aware attack metrics under matched evaluation protocols. The central finding is that fraud robustness is highly protocol-sensitive. Post-hoc filtering measures the filterability of unconstrained attacks, not the full success of a deployment-aware constrained attacker. On LCLD, under the tested CAPGD setting, integrating the instalment constraint and attacker mutability into attack generation raises feasible-flipped attacks from 3.7 under post-hoc filtering to 2,832.3 under projection plus mutability masking. The results on IEEE-CIS and Sparkov further show that one-hot validity is a recurring binding constraint, while Square Attack shows that protocol choice can change model-family conclusions.

Overall, financial fraud robustness evaluation should report predictive degradation and attack feasibility together. Domain constraints and attacker capability should be treated as part of attack generation, not merely as post-processing checks. Robust PR-AUC, aggregate feasibility, feasible-flipped count, and filtered success rate should be reported jointly.

\balance
\bibliographystyle{IEEEtran}
\bibliography{fraudbench_icdm2026_refs}

\end{document}